\pdfoutput=1

\documentclass[11pt]{article}

\usepackage[final]{acl}

\usepackage{times}
\usepackage{latexsym}
\usepackage{booktabs}
\usepackage{subcaption}
\usepackage{algorithm}
\usepackage{algpseudocode}
\algrenewcommand\alglinenumber[1]{\footnotesize #1:\ }
\usepackage{textcomp}
\usepackage{xcolor}
\usepackage{amsmath,amssymb,amsfonts}
\usepackage{graphicx}
\usepackage{algpseudocode}
\usepackage{float}
\usepackage{geometry}
\usepackage{pdflscape}
\usepackage{booktabs}
\usepackage{multirow}
\usepackage{hyperref}
\usepackage{tcolorbox}
\usepackage{amsmath}
\usepackage{graphicx} 
\usepackage{listings}
\usepackage[T1]{fontenc}

\usepackage[utf8]{inputenc}

\usepackage{microtype}

\usepackage{inconsolata}

\usepackage{graphicx}
\graphicspath{ {./latex/} }

\title{Momentum-Aided Natural Language Gradient Descent for Prompt Optimization}

\author{%
\textbf{Anthony Cui}\thanks{Lead Author} \quad \textbf{Pranav Nandyalam} \quad \textbf{Andrew Rufail} \quad \textbf{Ethan Cheung}
\\
\textbf{Aiden Lei} \quad \textbf{Kevin Zhu} \quad \textbf{Sean O'Brien}
\\
Algoverse AI Research\\
\texttt{anthonycui@u.northwestern.edu, kevin@algoverse.us}
}

\begin{document}
\maketitle
\begin{abstract}
Prompt optimization is crucial for improving the output quality of Large Language Models (LLMs), but many existing methods are inefficient, requiring extensive computation and manual tuning. We propose \textbf{M}omentum-\textbf{A}ided \textbf{P}rompt \textbf{O}ptimization (MAPO), which builds on ProTeGi \citep{pryzant} by incorporating positive natural language "gradients" and a momentum-based memory mechanism to refine prompts while avoiding local minima and oscillations. It also employs beam search and an Upper Confidence Bound (UCB) algorithm for balanced candidate expansion and selection. MAPO achieves faster convergence time with fewer API calls and higher performance than ProTeGi, demonstrating its effectiveness as a robust and scalable solution for automated prompt optimization in LLMs. Our code is available online at \href{https://github.com/AnthonyCui7/momentum-aided-prompt-optimization}{https://github.com/AnthonyCui7/momentum-aided-prompt-optimization}.

\end{abstract}

\section{Introduction}

Large Language Models (LLMs) have gained significant attention since the release of ChatGPT \cite{openai_chatgpt}, leading to the development of new prompting techniques that have greatly improved LLM performance \citep{schulhoff}. While  it has been shown that the prompts given to the LLM greatly affect performance \citep{pawlik2025choice}, prompts can still be unclear, biased, or incomplete, limiting LLM capabilities \citep{sahoo}. For these reasons, prompt engineering has become a critical aspect of leveraging an LLM’s capabilities. Often, current prompt engineering methods require manual adjustments by the user, making them time-consuming, error-prone, and constrained by human limitations \citep{lin}. This highlights an increasing need for an automated system to improve the quality of prompts without the need for human intervention.

\begin{figure}[h]
\centering
\includegraphics[scale=0.48]{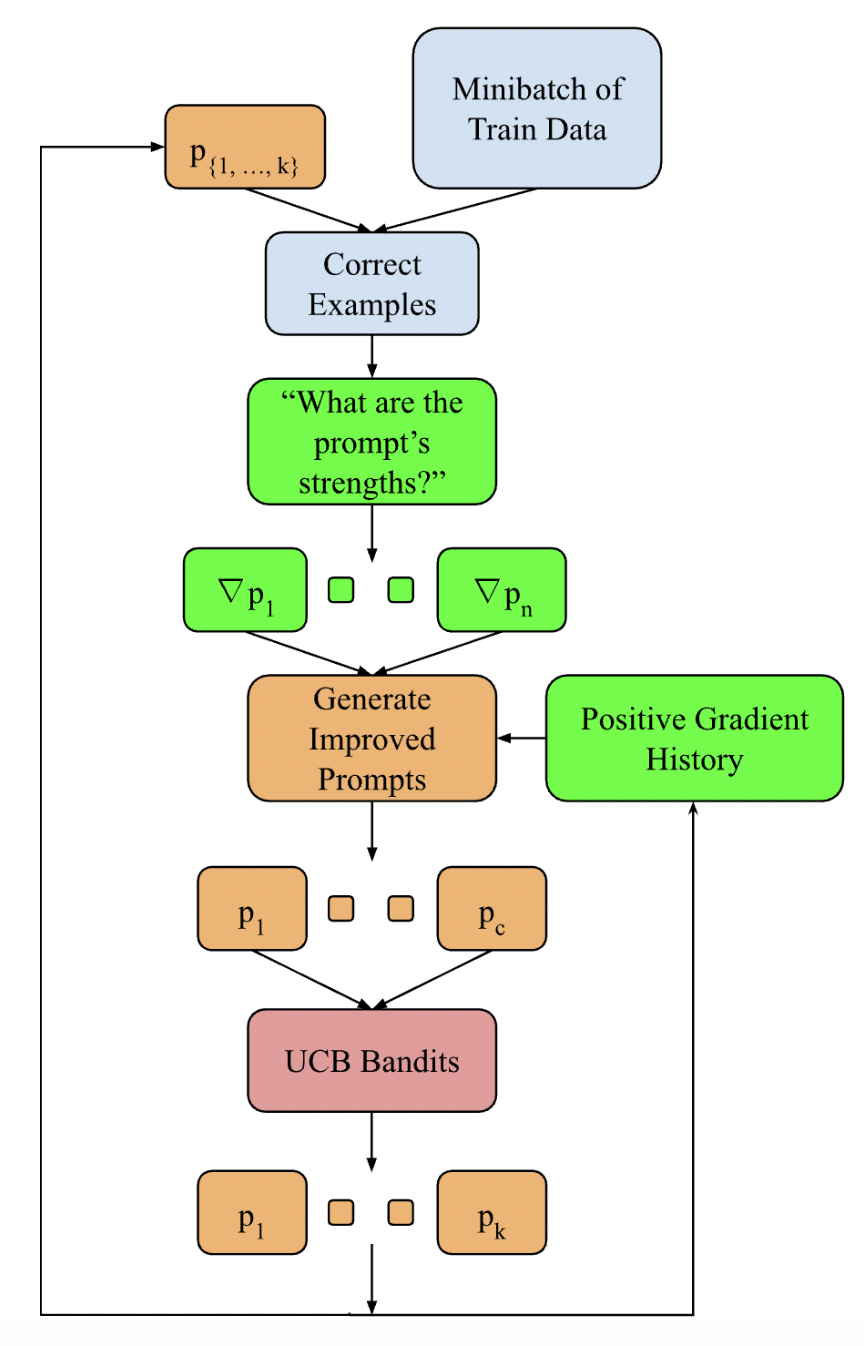}
\caption{High-Level Overview of MAPO}
\end{figure}

Recent work has explored implementing traditional machine learning algorithms in a natural language format, with one of the first being ProTeGi’s “Automatic Prompt Optimization with ‘Gradient Descent’ and Beam Search” \citep{pryzant}. While ProTeGi introduced an innovative framework, it has limitations, including high computational costs and resource consumption due to excessive API calls making large-scale prompt optimization impractical. Furthermore, ProTeGi does not track previous refinements, leading to oscillatory behavior and slow convergence. Ultimately, the strengths of prompts are underutilized.

We introduce Momentum-Aided Prompt Optimization (MAPO), a method that extends ProTeGi by using positive natural language “gradients” with momentum to automate prompt refinement. Gradients are suggested feedback for the current prompt generated using correct examples from a minibatch of training data, guiding the LLM to refine prompts in a consistent semantic direction. Beam search expands the generated candidate pool of improved child prompts, and the best-arm identification algorithm using UCB bandits, selects the top prompts for further evaluation. MAPO significantly improves upon ProTeGi by enhancing efficiency and effectiveness through momentum-based adjustments. Unlike ProTeGi, which treats each optimization step as an independent improvement, MAPO tracks the history of gradient updates, preventing oscillations and reducing the risk of suboptimal prompts (local minima in the semantic space). In our experimental results in \hyperref[sec:experiments]{Section 4}, MAPO achieves a 77.9\% reduction in convergence time compared to ProTeGi, 88.0\% reduction in convergence API calls, as well as a 5.28\% increase in peak performance, providing a scalable solution for automated prompt engineering in LLMs.

\section{Related Works}

\textbf{Prompt Optimization.} In this work, we draw upon existing prompt engineering techniques and focus on incorporating optimization algorithms into our framework to enhance the effectiveness of prompt optimization. There is an increasingly diverse set of general frameworks that previous works have focused on: LLM optimization \citep{pryzant, zelikman, fernando, zhou, yang}, reinforcement learning \citep{ma, zhang, deng}, and in-context learning \citep{shum}. However, these approaches are often infeasible when there is no architectural information introduced and only an API is provided to the LLM. More specifically, we base most of our work on improving automatic prompt engineering techniques with LLM gradient-based methods \citep{shin, pryzant}. Though, many of the current methods involving some form of iterative refinement technique, such as APE \citep{zhou} and ProTeGi all face similar struggles with cumulative costs of running their programs.

\section{Methods}
\subsection{Momentum-Aided Prompt Optimization}
First, the current prompt \( \mathbf{p} \) is evaluated based on a minibatch of training data to obtain randomly sampled strings \( \mathbf{s} \) representing correct LLM predictions aligned with ground-truth labels. We then provide the LLM with a static prompt \( \tau \) (see \autoref{sec:appendix-prompts}) to generate numerous positive “gradients” \( \nabla \mathbf{p} \) in natural language, appraising the current prompt \( \mathbf{p} \) using the sampled strings \( \mathbf{s} \). Our “gradients” \( \nabla \mathbf{p} \) are the natural language outputs of the LLM's continuation of static prompt \( \tau \). In traditional machine learning, gradient descent uses numerical gradients, representing a vector in parameter space that points in the direction of steepest ascent of the loss function, which the algorithm counters by moving in the opposite direction to minimize the loss function and improve model performance; in contrast, our natural language gradients \( \nabla \mathbf{p} \) represent directions in semantic space \citep{pryzant}. We use another static prompt \( \alpha \) (\autoref{sec:appendix-prompts}) to apply these gradients to the initial prompt \( \mathbf{p} \), allowing us to move along the same semantic direction as the positive natural language gradients, refining and improving the initial prompt.

\subsection{Expansion and Selection}

\begin{algorithm}
\caption{MAPO Beam Search}
\begin{algorithmic}[1]
    \Require $p_0$: initial prompt, $b$: beam width, $r$: search depth, $m$: metric function, $G$: gradient history
    \State $B_0 \gets \{p_0\}$
    \State $G_0 \gets \emptyset$
    \For{$i \gets 0$ \textbf{to} $r-1$}
        \State $C \gets \emptyset$
        \ForAll{$p \in B_i$}
            \State $C \gets C \cup \textit{Expand}(p, G_i)$ \Comment{Section 3.2}
        \EndFor
        \State $B_{i+1} \gets \textit{Select}_b(C, m)$ \Comment{UCB Bandits}
        \State $G_{i+1} \gets \textit{SampleGradient}(B_{i+1})$
    \EndFor
    \State $\hat{p} \gets \textit{argmax}_{p \in B_r}~ m(p)$
    \State \Return $\hat{p}$
\end{algorithmic}
\end{algorithm}

As outlined in Algorithm 1, our method employs beam search to explore the space of prompt variations generated during optimization. In each optimization round, the LLM utilizes static prompt \( \alpha \) containing sampled strings \( \mathbf{s} \) and gradients \( \nabla \mathbf{p} \) to edit the top \( k \) best-performing prompts from the previous round, producing $c$ new candidate prompts, denoted as \( \mathbf{p}_{c} \). After each round, less promising candidates are pruned, and only the top \( k \) prompts are retained for further gradient-based improvements, evaluated by a scoring function that assesses how well they meet our predefined objectives, such as the F1 score  \citep{manning2008introduction}.

In our implementation, we utilize an Upper Confidence Bound (UCB) Bandit Selection algorithm, which is applied during each beam search expansion to evaluate candidate prompts. The UCB Bandit Selection algorithm we employ is outlined in Algorithm 2 \citep{pryzant}.

\begin{algorithm}
\caption{UCB Bandits}
\begin{algorithmic}[1]
    \Require $n$ prompts $p_1, \dots, p_n$, dataset $\mathcal{D}_{\mathrm{tr}}$, $T$ time steps, metric function $m$, exploration $c_v$
    \State Initialize $N(p_i)\gets 0,\;Q(p_i)\gets 0\quad\forall i\in[1..n]$
    \For{$t = 1, \dots, T$}
        \State Sample $\mathcal{D}_{\text{sample}}\subset \mathcal{D}_{\text{tr}}$ uniformly
        \State $p_i \gets \arg\max_p \left\{ Q_t(p) + c_v \sqrt{\frac{\log t}{N_t(p)}} \right\}$ 
        \State Observe reward $r_{i,t} = m(p_i, \mathcal{D}_{\text{sample}})$
        \State $N_t(p_i) \gets N_t(p_i) + |\mathcal{D}_{\text{sample}}|$
        \State $Q_t(p_i) \gets Q_t(p_i) + \frac{r_{i,t}}{N_t(p_i)}$
    \EndFor
    \State \Return $\textsc{SelectTop}_b(Q_T)$
\end{algorithmic}
\end{algorithm}

We also record the gradients from static prompt \( \alpha \) used to generate the top \( k \) candidates in each round to incorporate our novel momentum extension into natural language gradient descent. Drawing on the physics intuition of momentum, traditional gradient descent uses this extension to improve stability and convergence, helping the model avoid oscillations and escape local minima, thereby reaching global minima more efficiently \citep{malingan}. Analogously, our method maintains a history of past gradients, guiding the movement of the initial prompt \( \mathbf{p} \) in each beam search round through semantic space, helping it converge on the optimal prompt rather than just an incremental improvement. A single positive gradient is randomly sampled from a pool of all the gradients used to generate the top \( k \) prompts in each beam search round, representing the positive gradient history. This gradient history is then incorporated into our static prompts \( \tau \) and \( \alpha \) as textual momentum, guiding the LLM to generate new gradients \( \nabla \mathbf{p} \) and new prompt candidates \( \mathbf{p}_{c} \), allowing the initial prompt \( \mathbf{p} \) to “roll down the hill” faster, i.e., achieve a much faster rate of convergence during prompt optimization.

\section{Experiments}
\label{sec:experiments}

\subsection{Setup}

Our experimental setup closely follows that of ProTeGi, allowing for a direct comparison between our extension method and their baseline. We use 200 randomly sampled data points as the test set, retaining most hyperparameters from ProTeGi's configuration, including a temperature of 0, a minibatch size of 64, and 6 rounds of beam search with a beam width of 4. Each parent prompt expands into 8 new candidate prompts, with 2 positive gradients generated from 3 randomly sampled correct examples from the minibatch, which creates fewer gradients than ProTeGi's method of using 4 negative gradients. This trade-off improves runtime efficiency while handling the added complexity of gradient history. The primary evaluation metric is the F1 score, and results reflect the highest score among the top $k$ beam search candidates, averaged over three trial runs. All experiments use the October 2024 release of GPT-3.5-turbo unless otherwise stated.

\begin{figure*}[h] 
    \centering
    \includegraphics[width=1\textwidth]{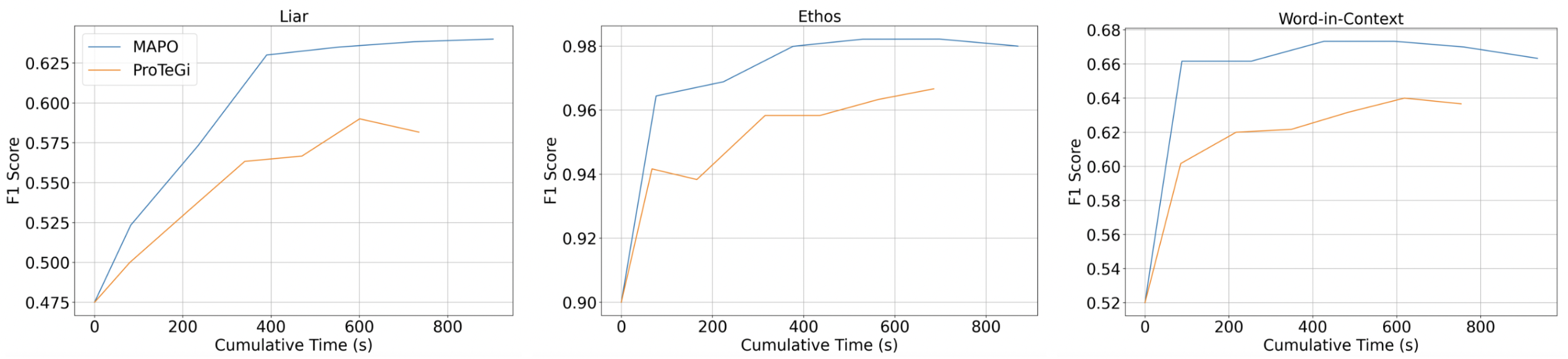}
    \caption{Test performance (F1 Score) versus Time for Liar, Ethos, and Word-in-Context dataset.}
    \label{fig:figure2}
\end{figure*}

\begin{figure*}[h] 
    \centering
    \includegraphics[width=1\textwidth]{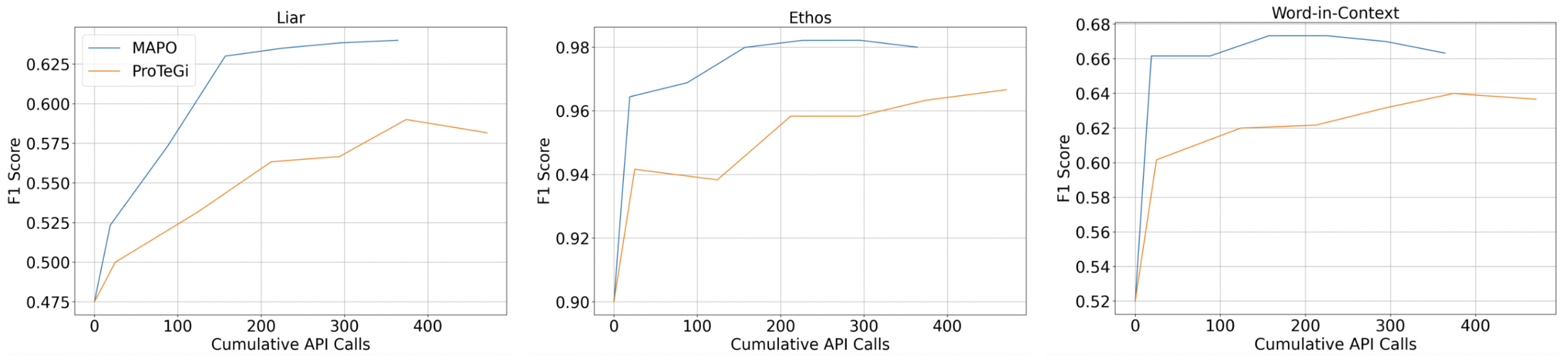}
    \caption{Test performance (F1 Score) versus Cumulative API Calls for Liar, Ethos, and Word-in-Context dataset.}
    \label{fig:figure3}
\end{figure*}

\begin{figure*}[h] 
    \centering
    \includegraphics[width=1\textwidth]{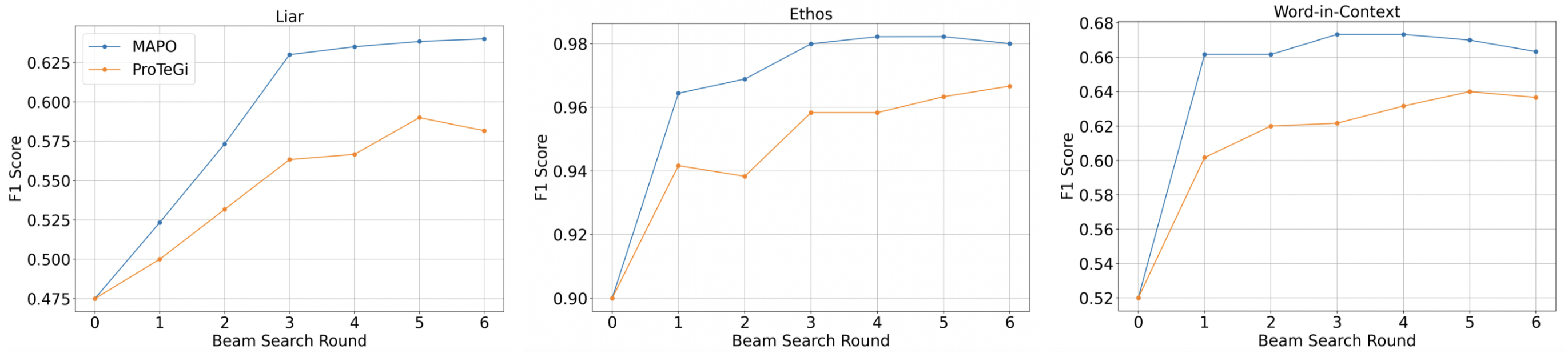}
    \caption{Test performance (F1 Score) versus Beam Search Round for Liar, Ethos, and Word-in-Context datasets.}
    \label{fig:figure4}
\end{figure*}

\begin{table*}[h]
\centering
\footnotesize
\begin{tabular}{llcc}
\toprule
\textbf{Dataset} & \textbf{Method/Metric} & \textbf{GPT-3.5-Turbo} & \textbf{GPT-4o Mini} \\
\midrule
\multirow{3}{*}{Ethos} 
    & \textit{Initial F1}   & 0.9 & 0.975 \\
    & MAPO                 & \textbf{0.98} / 871 / \textbf{364} / 146 / 56 / 2 & 0.985 / \textbf{1325} / \textbf{364} / 96 / 19 / 1 \\
    & ProTeGi              & 0.965 / \textbf{687} / 472 / - / - / - & 0.985 / 1328 / 423 / - / - / - \\
\midrule
\multirow{3}{*}{Liar} 
    & \textit{Initial F1}   & 0.475 & 0.52 \\
    & MAPO                 & \textbf{0.64} / 904 / \textbf{364} / 260 / 98 / 3 & 0.625 / 1414 / \textbf{364} / - / - / - \\
    & ProTeGi              & 0.58 / \textbf{737} / 472 / - / - / - & \textbf{0.64} / \textbf{1362} / 399 / - / - / - \\
\midrule
\multirow{3}{*}{WiC} 
    & \textit{Initial F1}   & 0.52 & 0.665 \\
    & MAPO                 & \textbf{0.66} / 937 / \textbf{364} / 73 / 16 / 1 & 0.73 / 1252 / \textbf{364} / 533 / 157 / 3 \\
    & ProTeGi              & 0.635 / \textbf{755} / 472 / - / - / - & 0.73 / \textbf{1119} / 406 / - / - / - \\
\bottomrule
\end{tabular}
\caption{Comparison of MAPO (ProTeGi) and baselines across tasks and models. Each cell shows: Final F1 Score / Total Time (s) / Total API Calls / Convergence Time (s) / Convergence API Calls / Convergence Optimization Steps. The row \textit{Initial F1} reports performance of the initial prompt before optimization.}
\label{tab:model_comparison}
\end{table*}

\begin{table*}[h]
\centering
\begin{tabular}{@{}lccc@{}} 
\toprule
\multicolumn{4}{c}{\textbf{Liar Dataset}} \\
\midrule
 & F1 Score & Conv. Time (s) & Total Time (s) \\
\cmidrule(l){2-4}
Positive Only (MAPO) & 0.66 & 285 & 903 \\
Negative Only & 0.60 & 303 & 1228 \\
Both Positive \& Negative & 0.65 & 275 & 1467 \\
\bottomrule
\end{tabular}
\caption{Comparison of Past Gradient Methods on the Liar Dataset. Convergence time represents time to reach peak ProTeGi performance for Liar}
\label{tab:past_methods}
\end{table*}

\subsubsection{Baseline}

\textbf{ProTeGi.} Developed by \cite{pryzant}, ProTeGi employs natural language gradient descent with negative gradients from incorrect example sampling to refine prompts. It iteratively applies these gradients to address prompt weaknesses, expanding the candidate pool using Monte Carlo sampling to generate paraphrased versions with synonyms or semantically similar variations. This ensures candidate diversity while guiding the optimization process.

\subsubsection{Benchmarks}

ProTeGi has been evaluated on benchmark datasets, such as the Liar Dataset \citep{wang}, which focuses on detecting fake news by classifying claims as true or false, and the Ethos Dataset \citep{mollas}, which is designed for identifying hate speech in online conversations. Our method will be evaluated on these same benchmarks for comparison. Additionally, we evaluate on the Word-in-Context dataset \citep{pilehvar}, which tests models on their ability to understand and represent the meanings of certain words based on the context in which they appear, to further expand the range of NLP tasks tested. Despite utilizing the publicly available code from ProTeGi to replicate the baseline, we were unable to reproduce the level of metric performance for their method on the Liar Dataset as reported in the original paper.

\subsection{Analysis}

\textbf{Efficiency.} \hyperref[fig:figure2]{Figures 2}, \ref{fig:figure3}, and \ref{fig:figure4} show the significant efficiency gains of MAPO over ProTeGi in terms of convergence time for MAPO to reach the same peak F1 score for ProTeGi. Overall, there is an average 77.9\% reduction in convergence time. This is notable since ProTeGi's runtimes can extend into hours \citep{pryzant}, highlighting the resource-intensive nature of automatic prompt optimization.

\autoref{fig:figure3} illustrates that ProTeGi uses about 417 API calls to reach its final F1 score, whereas MAPO requires an average of 56 to reach the same performance, resulting in an average 88.0\% reduction. MAPO also completes all six rounds of beam search with 105 fewer API calls on average, addressing critical computational constraints in large-scale optimization.

\autoref{fig:figure4} reinforces these efficiency gains by showing that MAPO surpasses ProTeGi's performance after just 2 or 3 optimization steps, while ProTeGi requires 6 steps for lower performance. This reduction in steps not only saves time but also suggests a more robust optimization mechanism in MAPO.

We attribute MAPO's efficiency primarily to its use of momentum, which allows it to incorporate feedback from previous optimization steps when refining prompts. Instead of treating each update independently, MAPO builds on prior successful changes, resulting in more consistent progress, fewer wasted iterations, and significantly faster convergence across multiple metrics. The use of positive gradients further constrains the search space to prompt edits supported by observed success, rather than encouraging arbitrary variation. Together, these design choices enable a more focused and informed optimization process, reducing the number of steps and API calls required to reach strong performance.

While MAPO does incur longer processing times per iteration due to more complex prompt structures and the inclusion of positive gradient history, the overall reduction in steps and runtime demonstrates that these trade-offs do not detract from its overall efficiency.

\begin{table}[h]
\centering
\begin{tabular}{lcc}
\toprule
\multicolumn{3}{c}{\textbf{Liar Dataset}} \\
\midrule
 & Time (s) & F1 Score \\
\cmidrule(l){2-3}
No Momentum & 390 & 0.64 \\
With Momentum & 285 & 0.64 \\
\midrule
\multicolumn{3}{c}{\textbf{Ethos Dataset}} \\
\midrule
 & Time (s) & F1 Score \\
\cmidrule(l){2-3}
No Momentum & 407 & 0.98 \\
With Momentum & 77 & 0.985 \\
\bottomrule
\end{tabular}
\caption{Comparison of Momentum Ablation on Liar and Ethos Datasets: Peak F1 Score and Convergence Time to Reach Peak ProTeGi Performance}
\label{tab:momentum_ablation}
\end{table}

\begin{table}[h]
\label{math}
\centering
\begin{tabular}{@{}lccc@{}} 
\toprule
\multicolumn{4}{c}{\textbf{Math Dataset Accuracy}} \\
\midrule
 & GSM8K&AGIEval-Math&\\
\cmidrule(l){2-4}
MAPO & \textbf{95.5}&\textbf{75.3}&\\
Original prompt & 94.5&74.1&  \\
\bottomrule
\end{tabular}
\caption{Comparison of MAPO's performance on math datasets. }
\label{tab:math}
\end{table}

\textbf{Efficacy.}  \autoref{fig:figure4} shows that MAPO consistently outperforms ProTeGi at every beam search round for both Liar and Ethos. MAPO's F1 score steadily increases while ProTeGi quickly converges and then plateaus or slightly declines. This results in a notable 5.28\% average increase in peak performance for MAPO.

MAPO's consistent improvement highlights its effectiveness in leveraging positive gradients and the momentum-based extension to thoroughly explore and refine the prompt search space. By incorporating momentum, MAPO helps it avoid local minima and erratic updates that hinder progress. This leads to a more robust optimization mechanism compared to ProTeGi.
In contrast, ProTeGi's performance dips suggest it struggles with optimization challenges like local minima and oscillations due to the lack of a stabilizing mechanism. MAPO's use of momentum and positive gradient history allows it to maintain steady progress, effectively "remembering" beneficial adjustments and reducing the likelihood of stagnation. This results in more reliable convergence toward higher performance levels, demonstrating the superior efficacy of MAPO's optimization strategy.

\textbf{Evaluation Across Models.}
Focusing on the performance of GPT-4o mini in \autoref{tab:model_comparison}, the model consistently demonstrates a strong initial F1 score, likely attributable to its advanced architecture. While ProTeGi yields competitive F1 scores with GPT-4o mini, MAPO achieves notably faster convergence in terms of optimization steps (1 or 3 steps) and convergence API calls (e.g., Ethos: 19 API calls for MAPO to reach ProTeGi performance with 423 API calls). This consistency of improved efficiency reinforces MAPO’s utility as a scalable and adaptable prompt optimization framework beyond the original GPT-3.5-Turbo setting.

\textbf{Gradient Ablation.}
As shown in \autoref{tab:past_methods}, the method using only positive gradients achieves the highest F1 score with the shortest total time, demonstrating its efficiency and effectiveness. The method combining both positive and negative gradients has the fastest convergence time, but its F1 score is slightly lower, and it requires the longest total time. This suggests that while the combined method offers faster convergence, it sacrifices some performance. On the other hand, the method using only negative gradients performs the worst across all metrics. These results highlight the importance of positive gradients for optimizing both accuracy and efficiency, while the combined method excels in rapid convergence scenarios. Although it takes longer to converge, the positive-only method significantly reduces overall runtime, as well as a slightly higher peak F1 score. Overall, we conclude that the benefits of a significantly shorter total time and a higher F1 score outweigh the slightly higher convergence time.

\textbf{Momentum Ablation.} 
In our momentum ablation study, we compared the convergence time required to reach peak ProTeGi performance with and without the use of momentum, as well as the peak MAPO F1 score performance. \autoref{tab:momentum_ablation} shows that the peak F1 score of MAPO remains essentially identical regardless of the inclusion of momentum, indicating that momentum does not compromise model accuracy.

However, there is a significant difference in convergence speed: on average, we observe a 54\% decrease in convergence time when incorporating momentum into MAPO. This substantial improvement in convergence speed demonstrates that momentum effectively enhances the optimization process by smoothing the loss landscape and helping the model avoid getting trapped in local minima, an issue that ProTeGi struggles with \cite{pryzant}. This is further evidenced by our smoother test set curves depicted in \autoref{fig:figure2}, \ref{fig:figure3}, and \ref{fig:figure4}, which contrast sharply with the oscillations observed in ProTeGi’s data. The lack of significant fluctuations in our graphs confirms that our method not only maximizes convergence speed but also promotes stability during training. Collectively, these findings validate the efficiency of incorporating momentum in prompt optimization, proving that our approach accelerates convergence while maintaining, or even slightly improving, the model's peak performance.

\textbf{Mathematical Reasoning.}
Since MAPO is a general prompt optimization algorithm, it can be easily generalized to many other tasks, not just binary classification. We test MAPO on the GSM8K dataset (Cobbe et al., 2021) which contains grade-level math word problems that some LLMs struggle with. We also test on the AGIEval-Math dataset (Zhong et al., 2023) which contains much harder questions. The results in \autoref{tab:math} are based on experiments using GPT-4o-mini and report accuracy as the evaluation metric, rather than F1 score.

The results presented in \autoref{tab:math} indicate that MAPO’s optimized prompt yields modest accuracy improvements in mathematical reasoning tasks. Specifically, it achieves a 1.2\% increase on AGIEval-Math and a 1.0\% increase on GSM8K.

\section{Conclusion}

In this work, we introduced Momentum-Aided Prompt Optimization (MAPO), a novel momentum-aided extension of natural language gradient descent for prompt optimization in LLMs. Building on ProTeGi, MAPO uses positive natural language gradients and momentum to refine prompts more effectively, guiding optimization consistently and avoiding local minima.

Momentum is crucial in overcoming ProTeGi's limitations such as local minima and oscillations. By leveraging gradient history, MAPO ensures a more stable, directed search, resulting in faster convergence and a more robust optimization process.

MAPO consistently outperforms ProTeGi, achieving a 77.9\% reduction in convergence time while improving peak F1 scores with fewer API calls and smoother convergence.

\section*{Limitations}
While MAPO demonstrates promising results, our evaluation remains limited in several key aspects. Firstly, we assessed MAPO on a narrow set of benchmarks and did not compare against a wide range of strong baselines, making it difficult to contextualize its performance relative to existing state-of-the-art methods. Future work should include a more extensive evaluation, including evaluations on diverse benchmarks and more comparative analysis.

\bibliography{custom.bib}

\begin{thebibliography}{20}
\providecommand{\natexlab}[1]{#1}

\bibitem[{Deng et~al.(2022)Deng, Wang, Hsieh, Wang, Guo, Shu, Song, Xing, and Hu}]{deng}
Mingkai Deng, Jianyu Wang, Cheng-Ping Hsieh, Yihan Wang, Han Guo, Tianmin Shu, Meng Song, Eric~P. Xing, and Zhiting Hu. 2022.
\newblock \href {https://arxiv.org/abs/2205.12548} {{RLPrompt: Optimizing Discrete Text Prompts with Reinforcement Learning}}.
\newblock \emph{arXiv.org}.

\bibitem[{Fernando et~al.(2023)Fernando, Banarse, Michalewski, Osindero, and Rocktäschel}]{fernando}
Chrisantha Fernando, Dylan Banarse, Henryk Michalewski, Simon Osindero, and Tim Rocktäschel. 2023.
\newblock \href {https://arxiv.org/abs/2309.16797} {{PromptBreeder: Self-Referential Self-Improvement via Prompt Evolution}}.
\newblock \emph{arXiv.org}.

\bibitem[{Lin et~al.(2024)Lin, Dai, Verma, Ng, Jaillet, and Low}]{lin}
Xiaohan Lin, Zhiqiang Dai, Anirudh Verma, Sufang Ng, Patrick Jaillet, and Bryan Kian~Hsiang Low. 2024.
\newblock \href {https://arxiv.org/abs/2405.17346} {{Prompt Optimization with Human Feedback}}.
\newblock \emph{arXiv.org}.

\bibitem[{Ma et~al.(2023)Ma, Liang, Wang, Huang, Bastani, Jayaraman, Zhu, Fan, and Anandkumar}]{ma}
Yecheng~Jason Ma, William Liang, Guanzhi Wang, De-An Huang, Osbert Bastani, Dinesh Jayaraman, Yuke Zhu, Linxi Fan, and Anima Anandkumar. 2023.
\newblock \href {https://arxiv.org/abs/2310.12931} {{Eureka: Human-Level Reward Design via Coding Large Language Models}}.
\newblock \emph{arXiv.org}.

\bibitem[{Malingan(2023)}]{malingan}
Navaneeth Malingan. 2023.
\newblock \href {https://www.scaler.com/topics/momentum-based-gradient-descent} {{Momentum-Based Gradient Descent - Scaler Topics}}.

\bibitem[{Manning et~al.(2008)Manning, Raghavan, and Sch{\"u}tze}]{manning2008introduction}
Christopher~D Manning, Prabhakar Raghavan, and Hinrich Sch{\"u}tze. 2008.
\newblock \emph{Introduction to Information Retrieval}.
\newblock Cambridge University Press.

\bibitem[{Mollas et~al.(2022)Mollas, Chrysopoulou, Karlos, and Tsoumakas}]{mollas}
Ioannis Mollas, Zoi Chrysopoulou, Sergios Karlos, and Grigorios Tsoumakas. 2022.
\newblock \href {https://doi.org/10.1007/s40747-021-00608-2} {{ETHOS: A Multi-Label Hate Speech Detection Dataset}}.
\newblock \emph{Complex \& Intelligent Systems}.

\bibitem[{{OpenAI}(2022)}]{openai_chatgpt}
{OpenAI}. 2022.
\newblock \href {https://openai.com/research/chatgpt} {{Introducing ChatGPT}}.
\newblock Accessed: 2024-10-19.

\bibitem[{Pawlik(2025)}]{pawlik2025choice}
Lukasz Pawlik. 2025.
\newblock How the choice of llm and prompt engineering affects chatbot effectiveness.
\newblock \emph{Electronics}, 14(5):888.

\bibitem[{Pilehvar and Camacho-Collados(2018)}]{pilehvar}
Mohammad~Taher Pilehvar and Jose Camacho-Collados. 2018.
\newblock \href {https://arxiv.org/abs/1808.09121} {{WiC: The Word-in-Context Dataset for Evaluating Context-Sensitive Meaning Representations}}.
\newblock \emph{arXiv.org}.

\bibitem[{Pryzant et~al.(2023)Pryzant, Iter, Li, Lee, Zhu, and Zeng}]{pryzant}
Reid Pryzant, Dan Iter, Jerry Li, Yin~Tat Lee, Chenguang Zhu, and Michael Zeng. 2023.
\newblock \href {https://arxiv.org/abs/2305.03495} {{Automatic Prompt Optimization with “Gradient Descent” and Beam Search}}.
\newblock \emph{arXiv.org}.

\bibitem[{Sahoo et~al.(2024)Sahoo, Singh, Saha, Jain, Mondal, and Chadha}]{sahoo}
Pallavi Sahoo, Ankit~Kumar Singh, Souvik Saha, Vipul Jain, Sourav Mondal, and Aditi Chadha. 2024.
\newblock \href {https://arxiv.org/abs/2402.07927} {{A Systematic Survey of Prompt Engineering in Large Language Models: Techniques and Applications}}.
\newblock \emph{arXiv.org}.

\bibitem[{Schulhoff et~al.(2024)Schulhoff, Ilie, Balepur, Kahadze, Liu, Si, Li, Gupta, Han, Schulhoff, Dulepet, Vidyadhara, Ki, Agrawal, Pham, Kroiz, Li, Tao, Srivastava, and Resnik}]{schulhoff}
Samuel Schulhoff, Madalina Ilie, Nihar Balepur, Kristine Kahadze, Alison Liu, Cheng Si, Yichen Li, Ananya Gupta, Hyejun Han, Samuel Schulhoff, Prashanth Dulepet, Suma Vidyadhara, Dong Ki, Saksham Agrawal, Christopher Pham, Guy Kroiz, Fangfei Li, Hao Tao, Aditya Srivastava, and Philip Resnik. 2024.
\newblock \href {https://arxiv.org/abs/2406.06608} {{The Prompt Report: A Systematic Survey of Prompting Techniques}}.
\newblock \emph{arXiv.org}.

\bibitem[{Shin et~al.(2020)Shin, Razeghi, IV, Wallace, and Singh}]{shin}
Taylor Shin, Yasaman Razeghi, Robert L~Logan IV, Eric Wallace, and Sameer Singh. 2020.
\newblock \href {https://arxiv.org/abs/2010.15980} {{AutoPrompt: Eliciting Knowledge from Language Models with Automatically Generated Prompts}}.
\newblock \emph{arXiv.org}.

\bibitem[{Shum et~al.(2023)Shum, Diao, and Zhang}]{shum}
KaShun Shum, Shizhe Diao, and Tong Zhang. 2023.
\newblock \href {https://arxiv.org/abs/2302.12822} {{Automatic Prompt Augmentation and Selection with Chain-of-Thought from Labeled Data}}.
\newblock \emph{arXiv.org}.

\bibitem[{Wang(2017)}]{wang}
William~Yang Wang. 2017.
\newblock \href {https://arxiv.org/abs/1705.00648} {{“Liar, Liar Pants on Fire”: A New Benchmark Dataset for Fake News Detection}}.
\newblock \emph{arXiv.org}.

\bibitem[{Yang et~al.(2023)Yang, Wang, Lu, Liu, Le, Zhou, and Chen}]{yang}
Chengrun Yang, Xuezhi Wang, Yifeng Lu, Hanxiao Liu, Quoc~V. Le, Denny Zhou, and Xinyun Chen. 2023.
\newblock \href {https://arxiv.org/abs/2309.03409} {{Large Language Models as Optimizers}}.
\newblock \emph{arXiv.org}.

\bibitem[{Zelikman et~al.(2023)Zelikman, Lorch, Mackey, and Kalai}]{zelikman}
Eric Zelikman, Eliana Lorch, Lester Mackey, and Adam~Tauman Kalai. 2023.
\newblock \href {https://arxiv.org/abs/2310.02304} {{Self-Taught Optimizer (STOP): Recursively Self-Improving Code Generation}}.
\newblock \emph{arXiv.org}.

\bibitem[{Zhang et~al.(2022)Zhang, Wang, Zhou, Schuurmans, and Gonzalez}]{zhang}
Tianjun Zhang, Xuezhi Wang, Denny Zhou, Dale Schuurmans, and Joseph~E. Gonzalez. 2022.
\newblock \href {https://arxiv.org/abs/2211.11890} {{TEMPERA: Test-Time Prompting via Reinforcement Learning}}.
\newblock \emph{arXiv.org}.

\bibitem[{Zhou et~al.(2022)Zhou, Muresanu, Han, Paster, Pitis, Chan, and Ba}]{zhou}
Yongchao Zhou, Andrei~Ioan Muresanu, Ziwen Han, Keiran Paster, Silviu Pitis, Harris Chan, and Jimmy Ba. 2022.
\newblock \href {https://arxiv.org/abs/2211.01910} {{Large Language Models are Human-Level Prompt Engineers}}.
\newblock \emph{arXiv.org}.

\end{thebibliography}

\nocite{Ando2005,andrew2007scalable,rasooli-tetrault-2015}

\appendix
\section{Prompts}
\label{sec:appendix-prompts}

MAPO uses two prompts. Prompt $\tau$ generates positive gradients based on correct examples sampled from the minibatch of training data, and prompt $\alpha$ applies these gradients to refine the current prompt and create child prompts.

\begin{lstlisting}
tau = f"""
        I'm trying to write a zero-shot {task_type} prompt.

        My current prompt is:
        "{prompt}"

        This prompt gets the following examples correct:
        {correct_string}

        In addition, consider the following strengths of past
        iterations of this prompt:
        {positive_gradient_history}

        Based on the above information, give {num_gradients} reasons why the 
        prompt could have gotten these examples correct.

        Wrap each reason with <START> and <END>
        """
\end{lstlisting}

\begin{lstlisting}
alpha = f"""
        I'm trying to write a zero-shot {task_type} solver.

        My current prompt is:
        "{prompt}"

        It gets the following examples correct:
        {correct_str}

        Based on these examples the strengths 
        with this current prompt are that {positive_feedback_str}

        Consider the following strengths
        of past iterations of this prompt:
        {positive_gradient_history}

        Based on the above information, 
        modify and revise the current prompt to create a new 
        prompt which improves upon the strengths of the 
        original wording.       
        The new prompt is wrapped with <START> and <END>.

        The 1 new prompt is:
        """    
\end{lstlisting}

\end{document}